\documentclass[conference,a4paper]{APSIPA2018}
\usepackage{multirow}
\usepackage{amsmath}
\usepackage[psamsfonts]{amssymb}
\usepackage{amsxtra}
\usepackage{graphicx,color}
\usepackage{threeparttable}
\usepackage{url}
\usepackage{epsfig,epstopdf}

\begin{document}

\title{AP18-OLR Challenge: Three Tasks and Their Baselines}

\author{%
\authorblockN{%
Zhiyuan Tang\authorrefmark{1},
Dong Wang\authorrefmark{1} and
Qing Chen\authorrefmark{2}
}
\authorblockA{%
\authorrefmark{1}
Tsinghua University}
\authorblockA{%
\authorrefmark{2}
SpeechOcean\\
Corresponding author: wangdong99@mails.tsinghua.edu.cn
}
}

\maketitle
\thispagestyle{empty}

\begin{abstract}

The third oriental language recognition (OLR) challenge AP18-OLR
is introduced in this paper, including the data profile, the tasks and
the evaluation principles.
Following the events in the last two years, namely AP16-OLR and AP17-OLR,
the challenge this year focuses on more challenging tasks, including
(1) short-duration utterances, (2) confusing languages,
and (3) open-set recognition.

The same as the previous events, the data of AP18-OLR is also
provided by SpeechOcean and the NSFC M2ASR project.
Baselines based on both the i-vector model and neural networks
are constructed for the participants' reference.
We report the baseline results on the three tasks and
demonstrate that the three tasks are truly challenging.
All the data is free for participants, and the Kaldi
recipes for the baselines have been published online.

\end{abstract}

\section{Introduction}

Oriental languages can be grouped into several language families,
such as Austroasiatic languages (e.g.,Vietnamese, Cambodia )~\cite{sidwell201114},
Tai-Kadai languages (e.g., Thai, Lao), Hmong-Mien languages (e.g., some dialects in south China), Sino-Tibetan languages (e.g., Chinese Mandarin), Altaic languages (e.g., Korea, Japanese) and Indo-European languages (e.g., Russian)~\cite{ramsey1987languages,shibatani1990languages,comrie1996russian}.
With the worldwide population movement and communication,
more and more multilingual phenomena are clear,
e.g., code switching between languages in an utterance
where the secondary languages may just appear as a single word.
The oriental languages themselves also influence each other via the multilingual interaction,
leading to complicated linguistic evolution. This complicated
multilingual phenomena attracted lots of research recently~\cite{huang2013cross,fer2015multilingual,wang2015transfer}.

To consistently boost the research on multilingual phenomena,
the center for speech and language technologies (CSLT) at
Tsinghua University and SpeechOcean organize the
oriental language recognition (OLR) challenge annually,
aiming at developing advanced language identification (LID) techniques.
The challenge has been conducted two times since 2016,
namely AP16-OLR~\cite{wang2016ap16} and AP17-OLR~\cite{tang2017ap17}.
They were very successful, especially AP17-OLR, in which
$31$ teams from $5$ countries participated.

AP17-OLR defined three test conditions
according to the duration of the test utterances:
$1$-second condition, $3$-second condition and full-utterance condition.
For the full-utterance condition,
the system submitted by the NUS-I2R-NTU team achieved the best performance
($C_{avg}$=$0.0034$, EER\%=$0.37$),
and for the $1$-second condition, the team SASI got the best performance
($C_{avg}$=$0.0765$, EER\%=$7.91$).
From these results, one can see that LID on long utterances have been solved
to a large extent, however for the short-utterance condition, the task remains
challenging. This is essentially the base when we designed the
OLR tasks this year. More details about the past two challenges can be found
on the challenge website.\footnote{http://www.olrchallenge.org}


Based on the experience of the last two years, we propose the third OLR challenge.
This new challenge, denoted by AP18-OLR, will be hosted by APSIPA ASC 2018.
It involves the same $10$ languages as in AP17-OLR, but focuses on more challenging
tasks: (1) short-duration utterances ($1$ second) LID, which inherits from AP17-OLR;
(2) LID for confusing language pairs; (3) open-set LID, where the test data
involves unknown interference languages.



In the rest of the paper, we will present the data profile and the evaluation plan of the AP18-OLR challenge. To
assist participants to build their own submissions, two types of baseline systems are constructed, based on the
i-vector model and various DNN models respectively. The Kaldi recipes of these baselines can be downloaded from the challenge web site.

\begin{table*}[htb]
\begin{center}
\caption{AP17-OL3 and AP16-OL7 Data Profile}
\label{tab:ol10}
\begin{tabular}{|l|l|c|c|c|c|c|c|c|}
 \hline
\multicolumn{3}{|c|}{\textbf{AP16-OL7}} & \multicolumn{3}{c|}{AP16-OL7-train/dev}  & \multicolumn{3}{c|}{AP16-OL7-test}\\
\hline
Code & Description & Channel & No. of Speakers & Utt./Spk. & Total Utt. & No. of Speakers & Utt./Spk. & Total Utt. \\
\hline
ct-cn & Cantonese in China Mainland and Hongkong & Mobile & 24 & 320 & 7559 & 6 & 300 & 1800 \\
\hline
zh-cn & Mandarin in China & Mobile & 24 & 300 & 7198        & 6 & 300 & 1800 \\
\hline
id-id & Indonesian in Indonesia &  Mobile & 24 & 320 & 7671 & 6 & 300 & 1800 \\
\hline
ja-jp & Japanese in Japan & Mobile & 24 & 320 & 7662        & 6 & 300 & 1800 \\
\hline
ru-ru & Russian in Russia & Mobile & 24 & 300 & 7190        & 6 & 300 & 1800 \\
\hline
ko-kr & Korean in Korea & Mobile & 24 & 300 & 7196          & 6 & 300 & 1800 \\
\hline
vi-vn & Vietnamese in Vietnam & Mobile & 24 & 300 & 7200    & 6 & 300 & 1800 \\
 \hline
\hline
 \multicolumn{3}{|c|}{\textbf{AP17-OL3}} & \multicolumn{3}{c|}{AP17-OL3-train/dev}  & \multicolumn{3}{c|}{AP17-OL3-test}\\
\hline
Code & Description & Channel & No. of Speakers & Utt./Spk. & Total Utt. & No. of Speakers & Utt./Spk. & Total Utt. \\
\hline
ka-cn & Kazakh in China & Mobile & 86 & 50  & 4200 &      86 &  20  & 1800 \\
\hline
ti-cn & Tibetan in China & Mobile & 34 & 330   & 11100 &    34 & 50  & 1800 \\
\hline
uy-id & Uyghur in China &  Mobile & 353 & 20   & 5800 &    353 & 5  & 1800 \\
 \hline
\end{tabular}
\begin{tablenotes}
\item[a] Male and Female speakers are balanced.
\item[b] The number of total utterances might be slightly smaller than expected, due to the quality check.
\end{tablenotes}
\end{center}
\end{table*}

\section{Database profile}

Participants of AP18-OLR can request the following datasets for system construction.
All these data can be used to train their submission systems.

\begin{itemize}
\item AP16-OL7: The standard database for AP16-OLR, including AP16-OL7-train, AP16-OL7-dev, and AP16-OL7-test.
\item AP17-OL3: A dataset provided by the M2ASR project, involving three new languages. It contains AP17-OL3-train and AP17-OL3-dev.
\item AP17-OLR-test: The standard test set for AP17-OLR. It contains AP17-OL7-test and AP17-OL3-test.
\item THCHS30:  The THCHS30 database (plus the accompanied resources) published by CSLT, Tsinghua University~\cite{wang2015thchs}.
\end{itemize}


Besides the speech signals, the AP16-OL7 and AP17-OL3 databases also provide lexicons of all the 10 languages, as well
as the transcriptions of all the training utterances. These resources allow training acoustic-based or phonetic-based
language recognition systems. Training phone-based speech recognition systems is also possible, though
large vocabulary recognition systems are not well supported, due to the lack of large-scale language models.

A test dataset will be provided at the date of result submission. This test set involves two parts: AP18-OL7-test and AP18-OL3-test.
The former involves utterances from the 7 target languages of AP16-OL7, but also 8 unknown interference languages.
The details of these databases are described as follows.

\subsection{AP16-OL7}

The AP16-OL7 database was originally created by Speechocean, targeting for various speech processing tasks.
It was provided as the standard training and test data in AP16-OLR.
The entire database involves 7 datasets, each in a particular language. The seven languages are:
Mandarin, Cantonese, Indonesian, Japanese, Russian, Korean and Vietnamese.
The data volume for each language is about $10$ hours of speech signals recorded in
reading style. The signals were
recorded by mobile phones, with a sampling rate of $16$ kHz  and a sample size of $16$ bits.

For Mandarin, Cantonese, Vietnamese and Indonesia, the recording was conducted in a quiet environment.
As for Russian, Korean and Japanese, there are $2$ recording sessions for each speaker: the first session
was recorded in a quiet environment and the second was recorded in a noisy environment.
The basic information of the AP16-OL7 database is presented in Table~\ref{tab:ol10},
and the details of the database can be found in the challenge website or the
description paper~\cite{wang2016ap16}.

\subsection{AP17-OL7-test}

The AP17-OL7 database is a dataset provided by SpeechOcean. This dataset contains 7 languages as in AP16-OL7,
each containing $1800$ utterances. The recording conditions are the same as AP16-OL7. This database is used as
part of the test set for the AP17-OLR challenge.

\subsection{AP17-OL3}

The AP17-OL3 database contains 3 languages: Kazakh, Tibetan and Uyghur, all are minority languages in China.
This database is part of the Multilingual Minorlingual Automatic Speech Recognition (M2ASR), which is
supported by the National Natural Science Foundation of China (NSFC). The project is a three-party collaboration, including Tsinghua University,
the Northwest National University, and Xinjiang University~\cite{wangm2asr}. The aim of this project is to construct speech recognition systems for five minor languages in China (Kazakh, Kirgiz, Mongolia, Tibetan and Uyghur). However, our ambition is beyond that scope: we hope
to construct a full set of linguistic and speech resources and tools for the five languages, and make them open and free for
research purposes. We call this the M2ASR Free Data Program. All the data resources, including the tools published in this paper, are released on the web site of the project.\footnote{http://m2asr.cslt.org}

The sentences of each language in AP17-OL3 are randomly selected from the original M2ASR corpus.
The data volume for each language in AP17-OL3 is about $10$ hours of speech signals
recorded in reading style.
The signals were recorded by mobile phones,
with a sampling rate of $16$ kHz and a sample size of $16$ bits.
We selected $1800$ utterances for each language as the development set (AP17-OL3-dev), and the rest is used as the
training set (AP17-OL3-train). The test set of each language involves $1800$ utterances, and is provided separately
and denoted by AP17-OL3-test.
Compared to AP16-OL7, AP17-OL3 contains much more variations in terms of recording conditions and
the number of speakers, which may inevitably  increase the difficulty of the challenge task.
The information of the AP17-OL3 database is summarized in Table~\ref{tab:ol10}.

\subsection{AP18-OLR-test}
The AP18-OLR-test database is the standard test set for AP18-OLR,
which contains AP18-OL7-test and AP18-OL3-test.
Like the AP17-OL7-test database,
AP18-OL7-test contains the same target $7$ languages, each containing $1800$ utterances,
while AP18-OL7-test also contains utterances
from several interference languages.
The recording conditions are the same as AP17-OL7-test.
Like the AP17-OL3-test database,
AP18-OL3-test contains the same $3$ languages, each containing $1800$ utterances.
The recording conditions are also the same as AP17-OL7-test.

\section{AP18-OLR challenge}

The evaluation plan of AP18-OLR keeps mostly the same
as in AP16-OLR and AP17-OLR, except some modification for the new
challenge tasks.

Following the definition of NIST LRE15~\cite{lre15}, the task of the LID challenge is defined
as follows: Given  a  segment  of  speech  and  a  language  hypothesis (i.e.,  a  target
language  of  interest  to  be  detected),  the  task  is  to decide  whether  that
target  language  was  in  fact  spoken  in  the given segment (yes or no), based on an
automated analysis of the data contained in the segment.
The evaluation plan mostly follows the principles of NIST LRE15.



The AP18-OLR challenge includes three tasks as follows:

\begin{itemize}
\item Task 1: Short-utterance identification task: This is a close-set identification task, 
   which means the language of each utterance is among the known $10$ target languages.
   The utterances are as short as $1$ second.
\item Task 2: Confusing-language identification task: This task identifies the language of utterances from 3 highly confusing languages (Cantonese, Korean and Mandarin).
\item Task 3: Open-set recognition task: In this task, the test utterance may be in none of the 10 target languages.
\end{itemize}

\subsection{System input/output}

The input to the LID system is a set of speech segments in unknown languages.
For task 1 and task 2, those speech segments are within
the $10$ known target languages,
while for task 3, the speech segment may be a non-target language.
The task of the LID system is to determine
the confidence that a language is contained in a speech segment. More specifically,
for each speech segment, the LID system outputs a score vector $<\ell_1, \ell_2, ..., \ell_{10}>$,
where $\ell_i$ represents the confidence that language $i$ is spoken in the speech segment.
The scores should be comparable across languages and segments.
This is consistent with
the principles of LRE15, but differs from that of LRE09~\cite{lre09} where an explicit decision
is required for each trial.

In summary, the output of an OLR submission will be a text file, where each line contains
a speech segment plus a score vector for this segment, e.g.,

\vspace{0.5cm}
\begin{tabular}{ccccccccc}
        & lang$_1$   & lang$_2$   & ... & lang$_9$  & lang$_{10}$\\
seg$_1$ & 0.5  & -0.2 &  ...& -0.3 & 0.1    \\
seg$_2$ & -0.1 & -0.3 &  ...& 0.5 & 0.3    \\
...   &      &     &  ... &      &
\end{tabular}

\subsection{Test condition}


\begin{itemize}
\item No additional training materials. The only resources that are allowed to use are: AP16-OL7, AP17-OL3, AP17-OLR-test and THCHS30.
\item All the trials should be processed. Scores of lost trials will be interpreted as -$\inf$.
\item The speech segments in each task
      should be processed independently, and each test segment in a group should be processed
      independently too. Knowledge from other test segments is not allowed to use (e.g.,
      score distribution of all the test segments).
\item Information of speakers is not allowed to use.
\item Listening to any speech segments is not allowed.
\end{itemize}

\subsection{Evaluation metrics}

As in LRE15, the AP18-OLR challenge chooses $C_{avg}$ as the principle evaluation metric.
First define the pair-wise loss that composes the missing and
false alarm probabilities for a particular target/non-target language pair:

\[
C(L_t, L_n)=P_{Target} P_{Miss}(L_t) + (1-P_{Target}) P_{FA}(L_t, L_n)
\]

\noindent where $L_t$ and $L_n$ are the target and non-target languages, respectively; $P_{Miss}$ and
$P_{FA}$ are the missing and false alarm probabilities, respectively. $P_{target}$ is the prior
probability for the target language, which is set to $0.5$ in the evaluation. Then the principle metric
$C_{avg}$ is defined as the average of the above pair-wise performance:


\[
 C_{avg} = \frac{1}{N} \sum_{L_t} \left\{
\begin{aligned}
  & \ P_{Target} \cdot P_{Miss}(L_t) \\
  &  + \sum_{L_n}\ P_{Non-Target} \cdot P_{FA}(L_t, L_n)\
\end{aligned}
\right\}
\]

\noindent where $N$ is the number of languages, and $P_{Non-Target}$ = $(1-P_{Target}) / (N -1 )$.
We have provided the evaluation script for system development.

\section{Baseline systems}

We constructed two kinds of baseline LID systems, based on the i-vector model and various DNN models respectively.
All the experiments were conducted with Kaldi~\cite{povey2011kaldi}.
The purpose of these experiments is to present a reference for the participants, rather than a competitive submission.
The recipes can be downloaded from the web page of the challenge.

\subsection{i-vector system}

The i-vector baseline systems were constructed based on the i-vector model~\cite{dehak2011front-end,dehak2011language}.
The static acoustic features involved 19-dimensional Mel
frequency cepstral coefficients (MFCCs) and the log energy.
This static features were augmented by their first and second order derivatives, resulting in 60-dimensional
feature vectors.
The UBM involved $2,048$ Gaussian components and the dimensionality of the i-vectors was $400$.
Linear discriminative analysis (LDA) was employed to promote language-related information.
The dimensionality of the LDA projection space was set to $150$.

With the i-vectors (either original or after LDA/PLDA transform), the score of a trail on a particular language
can be simply computed as the cosine distance between the test i-vector and the mean i-vector of
the training segments that belong to that language. This is denoted to
be `cosine distance scoring'.

\subsection{DNN systems}

For the DNN baseline, two kinds of DNN architectures were designed,
namely the traditional time-delay neural network (TDNN)~\cite{lang1990time} and
recurrent neural network with long short-term memory units (LSTM-RNN)~\cite{hochreiter1997long}.

The raw feature of all the two DNN systems is $40$-dimensional Fbanks,
with a symmetric $2$-frame window to splice neighboring frames.
For the TDNN LID, there are 6 hidden layers,
and the activation function is rectified linear unit (ReLU).
The number of units of each TDNN layer is set to be $650$.
The number of cells of the LSTM is set to be $512$.

\subsection{Performance results}

The primary evaluation metric in AP18-OLR is $C_{avg}$. Besides that, we also present the performance
in terms of equal error rate (EER). These metrics evaluate
system performance from different perspectives, offering a whole picture of the verification/identification capability
of the tested system.  At present, the performance is evaluated on the development set
which is actually the AP17-OLR-test database.
We present the utterance-level C$_{avg}$ and EER results for the three tasks respectively.

\subsubsection{Short-utterance LID}

The first task identifies short-duration utterances.
AP17-OLR-test contains three subset sets with different durations
($1$ second, $3$ second and regular length).
Besides the performance of the baseline systems on the $1$ second condition,
we also report the performance on the regular length for reference.
The results of i-vector and DNN systems
are showed in Table~\ref{tab:results1}.
From the results, we find that short-duration utterances
are hard to recognize for both the i-vector system and the DNN systems,
while DNN systems show more robustness.

\begin{table}[htb]
\begin{center}
\caption{C$_{avg}$ and EER results on $1$ second and full-length conditions}
\label{tab:results1}
\begin{tabular}{|l|c|c|c|c|}
\hline
               & \multicolumn{2}{|c|}{1 second } &  \multicolumn{2}{c|}{Full-Length}\\
\hline
System        &  $C_{avg}$  &   EER\%  &  $C_{avg}$  &   EER\% \\
\hline
\hline
i-vector       &   0.1888  &  18.75   &   0.0578  &  5.92  \\
i-vector + LDA &   0.1784  &  18.04   &   0.0598  &  6.12  \\
i-vector + PLDA &  0.1746  &  17.51   &   0.0596  &  5.86  \\
\hline
\hline
TDNN          &   0.1282  &  14.04   &   0.1034  &  11.31  \\
LSTM          &   0.1452  &  15.92   &   0.1154  &  12.76  \\
\hline
\end{tabular}
\end{center}
\end{table}

\subsubsection{Confusing-language LID}

The second task focuses on languages that are hard to distinguish (confusing languages).
To find the most indistinguishable languages among the $10$ target ones,
we investigated all possible language pairs (totally $45$ pairs)
and selected the most confusing ones.
The experiments showed that
different LID systems perform differently on different language
pairs.

By analyzing the results of all the systems on all the language pairs,
we found Cantonese, Korean and Mandarin are the most difficult to discriminate
from each other by both the i-vector and the DNN systems.
The results on the pairs of the three languages are shown in
Table~\ref{tab:result-2-pair}.
Finally, we put the three languages together to form the confusing-language set,
and baseline performance is reported in Table~\ref{tab:results2-3}. The
test utterances are the full-length set in AP17-OLR-test (this will be also
the case in the official test set).

The results indicate that the three languages are truly confusing and difficult
to distinguish from each other. This seems not surprising, as they
are spoken by people in neighboring areas.
Additionally, acoustic analysis shows that the inspiration and aspiration
of the pronunciation of these three languages are similar, and social linguistic
analysis shows that these languages influence each other in a significant way.
This is quite obvious as Chinese and Cantonese share almost the same characters,
and Korean borrowed many characters from Chinese, known as `hanja'.


\begin{table}[htb]
\begin{center}
\caption{C$_{avg}$ and EER results on pairs of Cantonese (Ca), Korean(Kr) and Mandarin (zh)}
\label{tab:result-2-pair}
\begin{tabular}{|l|c|c|c|c|c|c|}
\hline
       & \multicolumn{2}{|c|}{Ca/Kr} &\multicolumn{2}{|c|}{Ca/zh} &\multicolumn{2}{|c|}{Kr/zh} \\
\hline
System        &  $C_{avg}$  &   EER\%  &  $C_{avg}$  &   EER\% &  $C_{avg}$  &   EER\%   \\
\hline
\hline
i-vector       &   0.1684  &  17.00  & 0.1381  &  13.90  &   0.1172  &  11.40      \\
i-vector + LDA &   0.1569  &  15.72  & 0.1753  &  17.07  &   0.1246  &  12.40  \\
i-vector + PLDA &  0.1584  &  16.07  & 0.1850  &  18.36  &  0.1219  &  12.19\\
\hline
\hline
TDNN          &   0.3478  &  36.98   & 0.4360  &  61.72 &   0.1663  &  16.72 \\
LSTM          &   0.3080  &  36.60   & 0.4513  &  66.10  &   0.2131  &  21.45 \\
\hline
\end{tabular}
\end{center}
\end{table}

\begin{table}[htb]
\begin{center}
\caption{C$_{avg}$ and EER results on Cantonese, Korean and Mandarin}
\label{tab:results2-3}
\begin{tabular}{|l|c|c|}
\hline
System        &  $C_{avg}$  &   EER\%  \\
\hline
\hline
i-vector       &   0.1446  &  14.13     \\
i-vector + LDA &   0.1550  &  15.20   \\
i-vector + PLDA &  0.1563  &  15.64   \\
\hline
\hline
TDNN          &   0.3752  &  37.86   \\
LSTM          &   0.3902  &  40.56   \\
\hline
\end{tabular}
\end{center}
\end{table}

\subsubsection{Open-set language recognition}

In this task, utterances from non-target languages will be added to the test set.
To have a quick glimpse of the influence of the interference languages,
we chose $1,264$ utterances in non-target languages,
and combine them with $3,000$ utterances randomly selected from AP17-OLR-test,
leading to $42,640$ trials where $12,640$ non-target ones are from the interference languages.
The results with and without interference languages are shown in
Table~\ref{tab:results3}. From the results,
it can be seen that interference utterance indeed impacts the performance of LID systems,
particularly in terms of EER.
In the official test set that will be released, there will be a significant proportion of
utterances of non-target languages.

\begin{table}[htb]
\begin{center}
\caption{C$_{avg}$ and EER results with and without interference languages}
\label{tab:results3}
\begin{tabular}{|l|c|c|c|c|}
\hline
       & \multicolumn{2}{|c|}{Without interference} &  \multicolumn{2}{c|}{With interference}\\
\hline
System        &  $C_{avg}$  &   EER\%  &  $C_{avg}$  &   EER\% \\
\hline
\hline
i-vector       &   0.0556  &  5.93   &   0.0596  &  7.27  \\
i-vector + LDA &   0.0613  &  6.17   &   0.0643  &  7.40  \\
i-vector + PLDA &  0.0563  &  5.73   &   0.0606  &  7.13  \\
\hline
\hline
TDNN          &   0.1024  &  11.33   &   0.1056  &  13.53  \\
LSTM          &   0.1125  &  12.77   &   0.1159  &  14.77  \\
\hline
\end{tabular}
\end{center}
\end{table}


\section{Conclusions}

We presented the data profile, task definitions and evaluation principles of the AP18-OLR challenge.
To assist participants to construct a reasonable starting system, we published two types of baseline
systems based on the i-vector model and various DNN models respectively.
We showed that the tasks defined by AP18-OLR are rather challenging and are worthy of careful study.
All the data resources are free for the participants, and the recipes of the baseline systems can
be freely downloaded.


\bibliographystyle{IEEEtran}
\bibliography{ole}

\end{document}